\documentclass[10pt,twocolumn,letterpaper]{article}

\usepackage{iccv}
\usepackage{times}
\usepackage{epsfig}
\usepackage{graphicx}
\usepackage{amsmath}
\usepackage{amssymb}
\usepackage[accsupp]{axessibility}

\usepackage[pagebackref=true,breaklinks=true,letterpaper=true,colorlinks,bookmarks=false]{hyperref}

\iccvfinalcopy 

\def\iccvPaperID{22} 

\ificcvfinal\fi

\begin{document}

\title{Context-VQA: Towards Context-Aware and Purposeful Visual Question Answering}

\author{Nandita Naik\\
Stanford University\\
{\tt\small nanditan@cs.stanford.edu}
\and
Christopher Potts\\
Stanford University\\
{\tt\small cgpotts@stanford.edu}
\and
Elisa Kreiss\\
Stanford University\\
{\tt\small ekreiss@stanford.edu}
}

\maketitle
\ificcvfinal\thispagestyle{empty}\fi

\begin{abstract}
   Visual question answering (VQA) has the potential to make the Internet more accessible in an interactive way, allowing people who cannot see images to ask questions about them. However, multiple studies have shown that people who are blind or have low-vision prefer image explanations that incorporate the context in which an image appears, yet current VQA datasets focus on images in isolation. We argue that VQA models will not fully succeed at meeting people's needs unless they take context into account. To further motivate and analyze the distinction between different contexts, we introduce Context-VQA\footnote{More details about the dataset and code are available at \url{https://github.com/nnaik39/context-vqa}.}, a VQA dataset that pairs images with contexts, specifically types of websites (e.g., a shopping website). We find that the types of questions vary systematically across contexts. For example, images presented in a travel context garner 2 times more “Where?”\ questions, and images on social media and news garner 2.8 and 1.8 times more “Who?”\ questions than the average. We also find that context effects are especially important when participants can't see the image. These results demonstrate that context affects the types of questions asked and that VQA models should be context-sensitive to better meet people's needs, especially in accessibility settings.
\end{abstract}

\section{Introduction}

\begin{figure}
\centering
\includegraphics[width=0.9\linewidth]{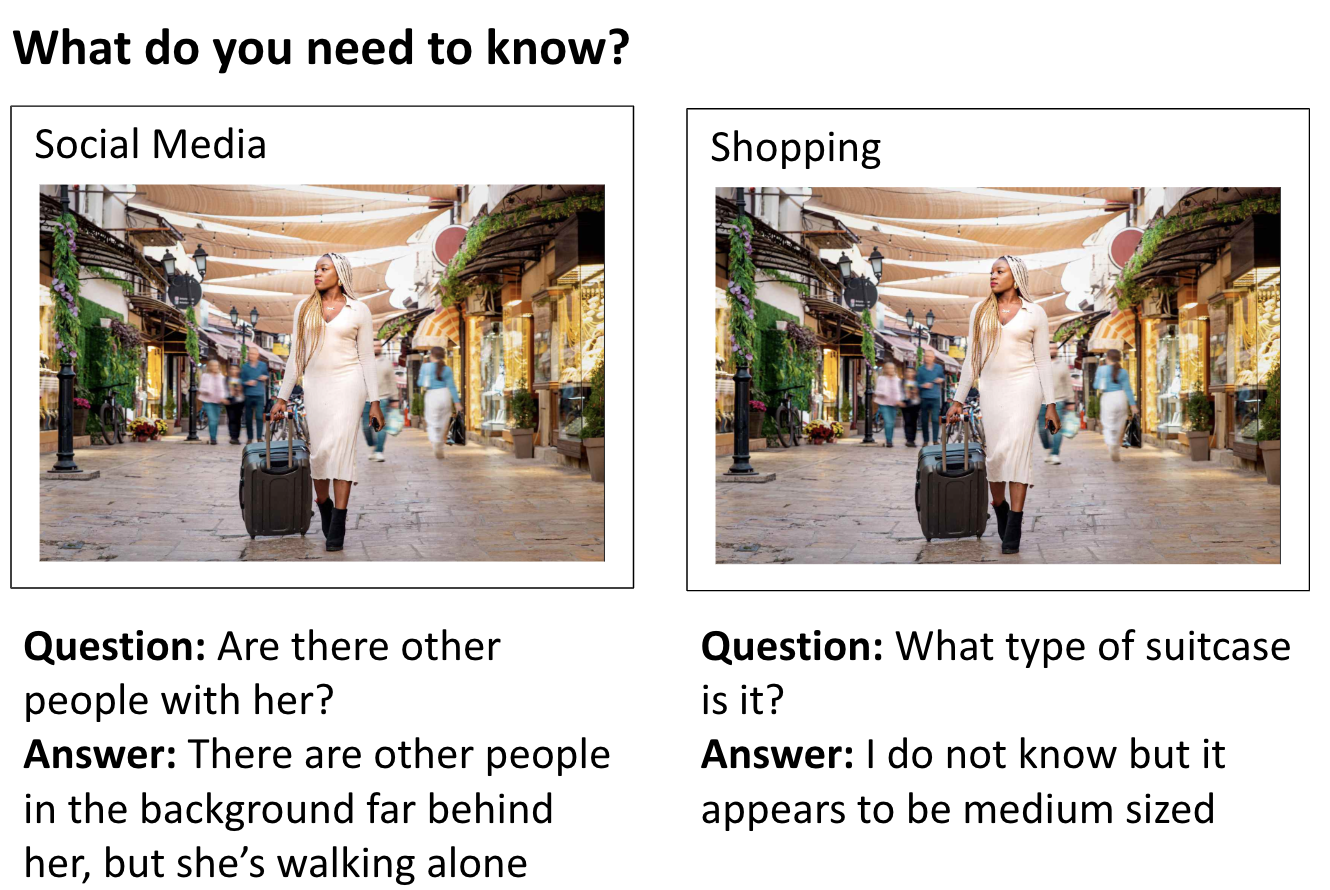}
\caption{We propose the dataset Context-VQA, where images are paired with various contexts. This example shows how both questions and answers vary across different contexts. On a social media website, the details about people within the image are more relevant, while in the context of a shopping website, people will want to know more specific details about the suitcase.}
\label{fig:suitcase_pic}
\end{figure}

Images are omnipresent on the Web, appearing in various contexts—for example, news websites, shopping websites, or social media \cite{image_search, describing_images_survey}. This poses an issue for nonvisual accessibility: How will people who cannot see the image (e.g., due to image loading issues or a visual impairment) understand the content that is presented? In this case, a textual interpretation of the image is needed. 
While much recent work has focused on leveraging AI to generate image descriptions \cite{dognin2022image, concadia}, being able to inquire about specific details gives the user agency over obtaining information that is specifically suited to their needs (which is, for example, leveraged in interactive accessibility studies \cite{muehlbradt2022s}). 
Visual Question Answering (VQA) models are a promising tool for providing image-based information on the fly. However, most VQA efforts are focused on evaluating machine ``understanding,'' an abstract task which seeks to test the spatial or object reasoning of machine learning models. 
The resulting systems aren't easily extendable to an accessibility application \cite{vqa_understanding_vs_accessibility}. The VizWiz dataset is an exception, where image--question pairs are obtained from the actual use case of blind and low-vision people asking questions about information in the visual world they're taking pictures of~\cite{vizwiz}. 

In contrast to traditional machine understanding VQA datasets, we propose Context-VQA, a dataset which recenters the focus on the use case of image accessibility and the changing information needs for images occurring in various sources online. Context-VQA complements VizWiz but is distinct from it, since the images are part of the general public discourse and not directly conditioned on a question. 

Current VQA tasks assume a one-size-fits-all approach, where each image is associated with a fixed set of questions. 
However, recent work suggests that different pieces of information become relevant depending on the particular context in which people encountered the image \cite{person_shoes_tree, beyondonesize}. Consider the question and answer provided in Figure \ref{fig:suitcase_pic}. It is difficult to infer the context from the image itself, yet the meaning of the image changes depending on the context. Our Context-VQA dataset presents images in distinct contexts to elicit more naturalistic question distributions. 

The purpose of the Context-VQA dataset is to align previous VQA efforts with an explicit accessibility goal, and to situate questions and answers within diverse Web contexts 
to more closely reflect user needs.
Each entry consists of a question, image, answer, and context, where the question and answer are both conditioned upon the context. We find that context shapes which questions become informative and that this contextual sensitivity is even more pronounced when participants are prompted to ask questions based on a description of the image and not the ground-truth image itself. This suggests that specifically in cases of image inaccessibility, the context plays a major role for VQA tasks.

Our contributions are:
\begin{enumerate}\setlength{\itemsep}{0pt}
\item A context-sensitive VQA dataset, with questions and answers that were written for images situated in specific Web contexts. To account for variation in question-writing strategies, half of the participants were shown the image and half were shown an AI-generated description of the image as a basis for their questions. The dataset therefore allows not only an investigation of context-sensitivity, but also of task formulation. While the image-visible version of the study might 
allow for more grounding, 
the description-only version imitates the user experience more closely.
\item An assessment of how Web context shapes which pieces of information in an image are of interest. We find that the distribution of question types varies across contexts in an intuitive manner. These findings underscore that the context affects the meaning of an image, and motivate context-aware, purposeful VQA models.
\end{enumerate}

\section{Related Work}

\paragraph{Context in Image Descriptions}
Making image content accessible through textual descriptions and explanations requires selecting pieces of visual information that appear most relevant. 
This has been a focus of investigation specifically for the usefulness of image descriptions. 
Traditionally, generating image descriptions takes a one-size-fits-all approach, where a single description is used across many different contexts. However, there is a growing consensus that people's information needs for the same image changes depending on the context where the image appears \cite{beyondonesize, person_shoes_tree,muehlbradt2022s,refless_metrics}. For example, while people prefer details about the color and attributes of the clothing on shopping websites, information about the relationships between people depicted in the image become more relevant in the context of social media. Furthermore, users also want image descriptions to clarify why the image appears on a site \cite{person_shoes_tree}. In an interactive image accessibility study between a BLV and a sighted user, this context dependence was also observable from the specific questions the BLV participant asked about the images \cite{muehlbradt2022s}, suggesting that context needs to play a role not just for image descriptions but also VQA systems. 

\paragraph{Visual Question Answering Datasets}
VQA tasks 
probe vision--language models for their alignment between the two domains. 
VQA-v2 \cite{vqa_v2} emphasizes open-ended, free-form questions. Visual7W \cite{visual7w} requires the model to ground its answer in specific regions of the image. CLEVR \cite{clevr} and GQA \cite{gqa} test spatial reasoning abilities. OK-VQA \cite{ok_vqa} requires outside knowledge to answer questions. These prior datasets are focused on evaluating machine ``understanding," an abstract task which tests spatial and logical reasoning \cite{vqa_understanding_vs_accessibility}. In contrast, Context-VQA seeks to prioritize the type of questions that are most relevant to \textit{people} based on contextually situated images.

The Visual Dialog \cite{visual_dialog} dataset established the task of ``visual chatbots," models that can hold a conversation about an image and answer follow-up questions. Similarly, the questions in Context-VQA were generated from giving study participants image descriptions and asking them to write follow-up questions. Where Context-VQA diverges from Visual Dialog, however, is the incorporation of context into the dataset, which moves away from a one-size-fits-all approach 
and towards reflecting the variety of questions asked for contextually situated images.

VizWiz \cite{vizwiz}, similarly to Context-VQA, is also designed to reflect image accessibility needs but for a distinct goal. 
All the images and the questions in VizWiz come from blind people who had a question about something in their environment, took a picture, and uploaded it to the VizWiz app along with their question, where crowdsourced human workers then provided an answer \cite{vizwiz}. For instance, someone might take an image of the error message on their rebooting computer screen or a clothing item of which they want to know the color. 
In VizWiz, the images are taken to explicitly answer a single question. However, the images that need to be made accessible online are often intended to meet multiple information needs in varied contexts.
Thus, 
Context-VQA addresses a separate image accessibility purpose than VizWiz and complements that effort.

\section{Dataset Construction}

The goal for the Context-VQA dataset is to allow for a direct comparison of how questions for images change based on 
image context.
To do so, we collected naturalistic images from six different types of websites, which were selected based on prior work \cite{beyondonesize}.
In a norming study, participants were presented with an image and, for each of the six contexts, rated how likely it was that the image appeared in the given context. We used this study to select plausible image-context pairs that were further annotated with questions (Experiment 1) and answers (Experiment 2). All human-subject experiments were run on the crowdsourcing platform Prolific 
under an IRB protocol.

\begin{figure*}
\begin{center}
\includegraphics[width=.9\linewidth]{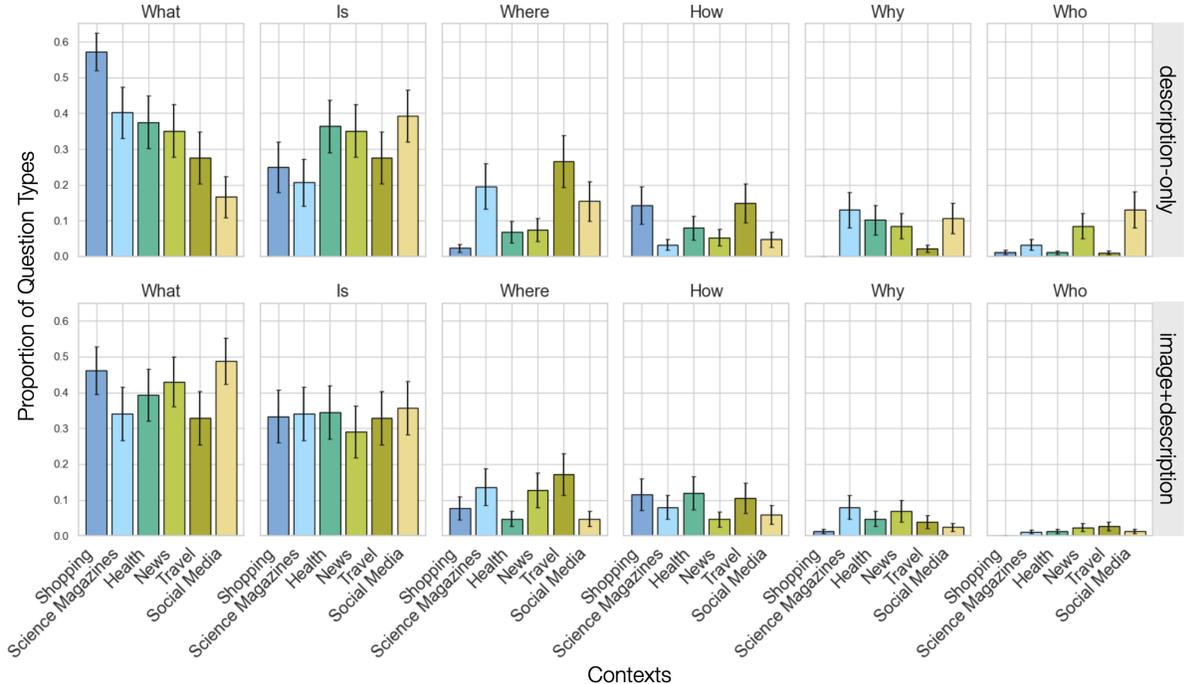}
\end{center}
\caption{Distribution of question types across contexts for the description-only study (top) and the image+description study (bottom), where error bars show variance.
The distinction between contexts is more pronounced in the description-only study, while in the image+description study, the question type proportions are more similar across contexts. Particularly in the description-only study, Social Media and News have more \emph{who} questions, Travel has more \emph{where} questions, and Shopping has more \emph{what} questions.}
\label{fig:question_type_across_contexts}
\end{figure*}

\subsection{Materials}

The contexts needed to be specific enough to be informative, yet broad enough to account for a range of images. Drawing inspiration from \cite{beyondonesize}, we selected six contexts: Shopping, Travel, Social Media, Health, News, and Science Magazines. These contexts also overlapped (e.g., an image of clothes could appear in Social Media and Shopping).

The images were sampled by hand, prioritizing high-quality digital images from diverse websites. Images were sourced in the Shopping category from clothes and dorm shopping websites; images in the Travel category from travel blogs and hotel websites; images in the Health category from health and fitness blogs; images on Social Media from Twitter, Pinterest, and Instagram; images in the Science category from Popular Mechanics and Popular Science; and images in the News category from the New York Times and San Francisco Chronicle. We selected seven images from each source context, resulting in 42 images. Based on the image-context norming study results, we selected 35 images for further annotation, which were rated to plausibly appear in up to 3 distinct contexts.

\subsection{Experiment 1: Question Generation}

Our first study aimed to generate questions conditioned on the image and context.

\textbf{Task}
There were two versions of the question generation task: description-only and image+description.
In the description-only task, participants were given a context and an image description generated by BLIP-2 \cite{blip2}. In the image+description version, participants were additionally shown the image itself. As an annotation quality check, we first asked them to write a justification for why the image might appear in the given context---low-effort or incomprehensible justifications resulted in data exclusion. They were then asked to suggest two follow-up questions that a person who only had access to the image description would ask. Prior work has noted that having the questioner not see the image could help remove visual priming bias--for example, questioners will often write, ``Is there a dog in this image?" only if there is a dog in the image \cite{visual_dialog}. The description-only condition also mimics the real-life condition, where people wouldn't see the image directly and ask follow-up questions based on a short description. 

We ran six trials. Each participant saw six different, randomly selected image--context pairs, and we ensured that no participant saw the same image or context repeated twice.

\textbf{Participants}
We recruited 55 participants over Prolific. They took an average of 8.5 minutes to complete the trials and were paid \$13.52 per hour on average.

\textbf{Post-processing} We collected 1,320 questions across both studies. We excluded participants who expressed in the post-questionnaire that they might have done the study incorrectly, or whose written justifications for why the image appeared in the context was less than 30 characters, resulting in 1,096 questions. We also manually filtered questions that were inappropriate or not questions, leaving 1,032 questions. For each of the 65 unique image-context pairs, we selected two questions at random from each study for further answer annotation.

\subsection{Experiment 2: Answer Generation}

For the questions generated in Experiment 1, we collected answer annotations.

\textbf{Task} 
Each participant saw six unique image--context pairs, associated with a question, which were randomly sampled from our dataset of questions. They were asked to justify the occurrence of the image in the context and answer the question based on the image. They also had an option to indicate if the question was unanswerable.

\textbf{Participants}
We recruited 100 participants over Prolific. Participants took an average of 8.5 minutes to complete and were paid an average of \$13.57 per hour.

\textbf{Post-processing}
We collected 1,260 answers for 202 questions. We excluded participants who expressed in the post-questionnaire that they might have done the study incorrectly, or whose written justifications for why the image appeared in the context was less than 30 characters. After exclusions, the total number of answers collected was 568, averaging 3 question-answers per image-context pair, 8 of which were rated unanswerable by at least 3 people.

\section{Dataset Analysis}

Based on prior research, we hypothesized that the context images are presented in shapes the questions participants will ask. To test this hypothesis, we investigated whether participants chose different question types dependent on the context. For analysis, we selected the question types of \emph{what}, \emph{who}, \emph{where}, \emph{why}, \emph{is}, \emph{how}, and \emph{when}, where \emph{is} questions represent all questions requiring only binary yes/no responses (i.e., \emph{are}, \emph{does}, \emph{was}, and \emph{is}). We prompted GPT-4 \cite{openai2023gpt4} to label all questions with their question type, and manually verified the labels. \emph{When} questions only occurred once, so we omitted them in the analysis. 

Figure \ref{fig:question_type_across_contexts} shows the question type distribution across contexts for all questions collected from the description-only study (top) and the image+description study (bottom).
For both study conditions, \emph{what} and binary response questions (\emph{is}) occur most frequently. We observe clear effects of context, especially in the description-only study. For instance, questions asked for images that appear in the shopping context are more likely to be \emph{what} questions ($p<0.001$) and less likely to be \emph{where} questions ($p<0.01$) compared to other contexts.\footnote{All statistical tests are generalized linear model analyses.} Questions for images appearing on social media, however, are the least likely to be \emph{what} questions ($p<0.001$) but are the most likely to be \emph{who} questions ($p<0.001$).
These results can help provide insight into where a model must be especially robust, and they highlight the importance of systems that can adjust to contextually changing user needs. 

While some context effects are reflected in the image+description study (e.g., in the \emph{where} and \emph{how} distributions), they are overall less pronounced than in the description-only study.
We hypothesize this is because the questioners have less information in the description-only study to inform their questions. These results suggest that context-sensitivity in VQA needs to play an especially important role when the image itself isn't visible to the end user, and is therefore particularly relevant for the image accessibility setting. Image+description VQA study setups might provide misleading results, since the presence of the image dampens the context effect.

Prior work suggests that when questioners don't see the image, they ask longer and more open-ended questions, which necessitates more detailed answers \cite{visual_dialog}. 
The average response length for our answers, across question types and study conditions, ended up being 11.03 words. This shows that the answers in our dataset are meaningful and not restricted to binary responses. This is an uncommonly long average response length for VQA datasets--for a point of comparison, Visual Dialog \cite{visual_dialog} reported an average answer length of 2.9 words, which was still higher than VQA \cite{vqa} (1.1 words), and Visual 7W \cite{visual7w} (2.0 words). We hypothesize the main contributors to answer length are expressions of uncertainty or additional details.
\section{Conclusion}

We present the Context-VQA dataset and use it to show how context changes the distribution of question types for VQA models. Unlike previous VQA datasets, Context-VQA is context-sensitive and reflects the distribution of question types that people will ask in different contexts. We also show initial evidence that the questions vary across contexts, especially when the image can't be seen.



{\small
\bibliographystyle{ieee_fullname}
\bibliography{egbib}
}

\end{document}